\documentclass[preprint,authoryear]{elsarticle}

\usepackage[utf8x]{inputenc} 

\usepackage{textcomp}
\usepackage{lmodern}
\usepackage{float}
\usepackage{multicol}

\usepackage{colortbl}
\usepackage[dvipsnames]{xcolor}
\usepackage{rotating}
\usepackage{placeins}

\usepackage{graphicx}
\usepackage{subfigure}
\usepackage{media9}

\usepackage{longtable}
\usepackage{multirow}
\usepackage{caption}

\usepackage{amsmath}
\usepackage{amssymb}
\usepackage{amsthm}
\usepackage{nameref}
\usepackage{wrapfig}
\usepackage{ulem}
\usepackage{float}
\usepackage{siunitx}

\usepackage[ruled,linesnumbered]{algorithm2e}

\usepackage{booktabs}	

\usepackage[breaklinks=true,colorlinks=false, pdfborder={0 0 0}]{hyperref}
\usepackage{color}

\usepackage{url}
\usepackage{epsfig}
\usepackage{placeins} 
\usepackage{longtable} 
\usepackage{marginnote} 

\DeclareMathAlphabet{\mathpzc}{OT1}{pzc}{m}{it}

\theoremstyle{plain}

\theoremstyle{definition}

\renewcommand{\digamma}[1]{\Psi\left(#1\right)}

\newcommand{\rev}{}

\newcommand{\mabbr}{AUDIT}
\newcommand{\mname}{\textbf{AU}tomated \textbf{DI}stance es\textbf{T}imation (\mabbr)}

\begin{document}

\title{\Huge\bfseries Automated Distance Estimation 
for Wildlife Camera Trapping}

\author[1]{Peter Johanns}
\ead{johanns@uni-bonn.de}
\author[1]{Timm Haucke}
\ead{haucke@cs.uni-bonn.de}
\author[1]{Volker Steinhage}
\ead{steinhage@cs.uni-bonn.de}

\date{\today}

\address[1]{University of Bonn, Institute of Computer Science IV, Friedrich-Hirzebruch-Allee 8, Bonn 53115, Germany}


\begin{abstract}
\rev{The ongoing biodiversity crisis calls for accurate estimation of animal density and abundance to identify sources of biodiversity decline and effectiveness of conservation interventions. Camera traps together with abundance estimation methods are often employed for this purpose. The necessary distances between camera and observed animals are traditionally derived in a laborious, fully manual or semi-automatic process. Both approaches require reference image material, which is both difficult to acquire and not available for existing datasets. We propose a fully automatic approach we call \mname{} to estimate camera-to-animal distances. We leverage existing state-of-the-art relative monocular depth estimation and combine it with a novel alignment procedure to estimate metric distances. \mabbr{} is fully automated and requires neither the comparison of observations in camera trap imagery with reference images nor capturing of reference image material at all. \mabbr{} therefore relieves biologists and ecologists from a significant workload.

We evaluate \mabbr{} on a zoo scenario dataset unseen during training where we achieve a mean absolute distance estimation error over all animal instances of only 0.9864 meters and mean relative error (REL) of 0.113.
The code and usage instructions are available at
{\small\href{https://github.com/PJ-cs/DistanceEstimationTracking}{https://github.com/PJ-cs/DistanceEstimationTracking}}


}%
\end{abstract}


\maketitle

\let\thefootnote\relax\footnotetext{
\\
© 2022. This manuscript version is made available under the CC-BY-NC-ND 4.0 license \href{https://creativecommons.org/licenses/by-nc-nd/4.0/}{https://creativecommons.org/licenses/by-nc-nd/4.0/}
}

\clearpage



\section{Introduction}

 
    
    
    

The biodiversity \rev{crisis requires} an accurate monitoring of animal density and abundance. Such estimates can then be used to identify causes of biodiversity loss and quantify the effects of conservation efforts. This is often achieved by employing camera traps, which capture image or video upon detection of an animal by a passive infrared sensor. Capture-recapture models can be used to estimate animal abundance by (re-)identifying individual animals over multiple images \citep{o2011camera}, which is however difficult for species without individual markings.

\rev{%
Within a joint project on building up a network of Automated Multisensor stations for Monitoring Of species Diversity (AMMOD) \citep{WAGELE2022105}, one sub-project is devoted to abundance estimation using Camera Trap Distance Sampling (CTDS) \citep{ctds}. 
Since CTDS relies on the manual and laborious evaluation of reference images, \citet{haucke2022overcoming} presented a first approach to overcome the distance estimation bottleneck in estimating animal abundance by proposing a semi-automatic calibration procedure. Experiments have shown that this semi-automated approach reduced the manual effort for calibration by reference images by a factor greater than 21. 
But the semi-automated approach is still requiring reference image material, which is both difficult to acquire and not available for existing datasets. In a proof-of-concept study, we propose now a fully automatic approach to estimate camera-to-animal distances and evaluate it in our CTDS framework.
}%

\subsection{Problem Statement \& Contributions}
\rev{
\noindent
From the application perspective, there are two obstacles to using automated approaches to abundance estimation efficiently:
\begin{itemize}
    \item \textbf{Demand for reference imagery}. Previous works require reference images \citep{ctds,haucke2022overcoming}, which are costly to obtain and often not available for existing datasets.
    \item \textbf{Demand for local placement of reference objects}. Although the process of manually comparing observation images with reference images has been automated, reference objects must still be located by hand \citep{haucke2022overcoming}.
\end{itemize}
Therefore, we show in a proof-of-concept study within our project-related CTDS framework a methodological new approach to \mname{} to overcome both obstacles (cf. Fig. \ref{fig:workflow}):
\begin{itemize}
    \item \textbf{No need at all for reference imagery} using a fully automated processing pipeline with a novel alignment procedure that is able to derive metric depth images that capture the absolute distances between camera traps and observed animals in meters from  the corresponding pure camera trap-based observation images.
    \item \textbf{No need at all for local placement of reference objects} since the observed animals themselves are automatically detected and localized with per-animal distance estimations.
\end{itemize}

} 

\rev{
\subsection{Related Work and Background}

\noindent
Here we present on the one hand the most relevant related work with respect to our proof-of-concept study. On the other hand, this goes along -- at least partially -- with a generally understandable concise introduction of essential methodological background and terminology that originates from the field of computer vision.

\subsubsection{Abundance Estimation}
\noindent
Several abundance estimation methods for unmarked animal populations have been proposed, which do not require the identification of individuals: the random encounter model (REM) \citep{rowcliffe2008estimating}, the random encounter and staying time model (REST) \citep{nakashima2018estimating}, the time-to-event model (TTE), space-to-event model (STE), instantaneous estimator (IS) \citep{moeller2018three} and camera trap distance sampling (CTDS) \citep{ctds}. While these methods do not require the reidentification of individual animals, they do require an estimation of the effective area surveyed by the camera trap. The effective surveyed area is dependent on the Field of View (FOV) of the camera trap and its effective detection distance. The effective detection distance is the distance below which as many individuals are missed as seen beyond \citep{hofmeester2017simple}. Estimating the effective detection distance generally requires estimating the distance between the camera trap and the detected animals. So far, three different approaches have been available to derive such camera-animal distances. However, they rely either on the manual and laborious evaluation of reference images \citep{ctds}, even more time-consuming on-site distance measurements \citep{camerasens} or semi-automatic calibration of relative depth images for the specific sequence \citep{haucke2022overcoming}.

\rev{While our proof-of-concept study focuses on our project-related application within the framework of camera trap
distance sampling (CTDS), we conjecture that our basic algorithms to estimate camera-to-animal distances transcend to the reported broader range of abundance estimation approaches.}

\subsubsection{Image-based distance estimation}
\noindent
Computer stereo vision is the traditional and well-established approach to image-based distance estimation. By comparing information about an observed scene from two differing camera perspectives (mostly two cameras, displaced horizontally from one another), depth information can be derived. \footnote{In this contribution, the two terms \textit{distance} and \textit{depth} refer in the same way to the distance between the camera and the points observed in a scene.} Computer stereo vision can be seen as the technical analogue to human stereopsis, that is, human perception of depth and three-dimensional structure by combining visual information from our two eyes.

However, currently all deployed camera traps do not utilize two cameras for stereo vision but use just one camera yielding one sight on the observed scene. This is called \textit{monocular vision}. 

Recent developments have shown that detailed distance estimations can be derived from images of conventional monocular cameras based on deep learning approaches \citep{Facil_2019_CVPR}. Meanwhile, various deep learning approaches have shown their effectiveness in addressing this so-called monocular depth estimation (MDE) where \textit{depth} is a synonym for the distance to the camera. In this way, monocular vision via deep learning can be seen as the technical analogue of a one-eyed human who learns to estimate distances by experience.  

MDE has a wide variety of applications, for example, augmented reality applications \citep{Woo2011DepthassistedR3}, real-time 3D human pose and movement estimation \citep{Moon_2018_CVPR}, navigation of autonomous vehicles \citep{kitti} or robots, 3D photography \citep{Kopf-OneShot-2020}, 3D scene reconstruction (Fig. \ref{S01_3D_tracks}) and many more. 

In this proof-of-concept study in the application framework of abundance estimation, we decide for the DPT (Dense Prediction Transformers) approach that has shown superior quantitative and qualitative results in MDE \citep{Ranftl2021}.

\rev{But most MDE approaches estimate only \textit{relative depth} information, where the distance-wise order and relative distances between objects in the scene are known (e.g., ``point A is closer to the camera than point B''), but not \textit{absolute depth} information in meters (e.g., ``points A and B show distances of 1.25 meter and 2.45 meter to the camera, respectively'') which is decidable for distance estimation in the framework of abundance estimation.
}
In this proof-of-concept study, we propose a novel alignment procedure to derive absolute depth information from relative depth information.

Relative depth information as well as absolute depth information will be visualized in this contribution by so-called \textit{\textbf{heatmaps}} where a color-based encoding depicts depth information (cf. Fig. \ref{fig:heatmaps}).

\begin{figure}
    \centering
    \includegraphics[height=3.0cm]{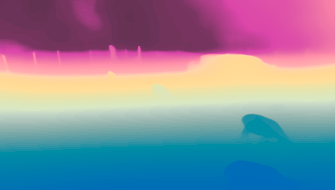}
    \hspace{0.25cm}
    \includegraphics[height=3.0cm]{Graphics/rel_depth_new.png}
    \includegraphics[height=3.0cm]{Graphics/rel_depth_new_colorbar.png}
    \caption{Left: A heatmap depicting relative depth information where distance is highest in pink and lowest in cyan. Right: A heatmap depicting absolute depth information where each color value corresponds to a metric distance to the camera.}
    \label{fig:heatmaps}
\end{figure}

\subsubsection{Visual animal detection}
\noindent
First so-called region-based deep learning approaches to visual object detection delivered for an input image for each detected object a so-called bounding box as output, where a bounding box is just a rectangle containing the detected object. Methods such as Mask R-CNN \citep{MaskRCNN} predict for each detected object not only a bounding box but also a so-called \textit{segmentation mask}. A segmentation mask shows the exact visual appearance of an detected object, that is, all pixels that belong to the visual appearance of an detected object (cf. Fig. \ref{fig:Mask&BB}). Segmentation masks of detected animals are important for behavioral studies of individual animals and animal herds based on their poses and actions captured by video clips from camera traps \citep{SCHINDLER2021101215}.

In this proof-of-concept study, we utilize the MegaDetector for visual animal detection in terms of bounding boxes. The MegaDetector which is an animal detection method for camera-trap footage developed by \citep{beery2019efficient} and was trained on several hundred thousand animal detections from camera trap videos recorded in diverse biospheres and of a large variety of animals. Based on the bounding boxes of detected animals, we introduce a new so-called \textit{multi-instance DINO} foreground segmentation to derive the segmentation masks of detected animals.

\begin{figure}
    \centering
    \includegraphics[height=3.0cm]{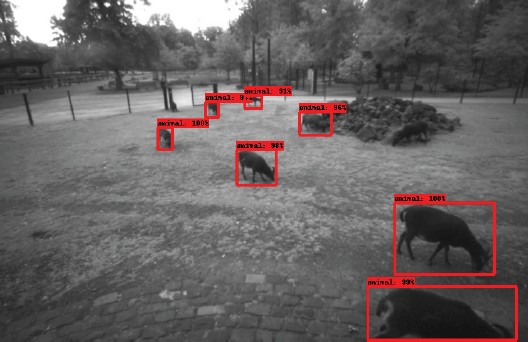}
    \hspace{0.25cm}
    \includegraphics[height=3.0cm]{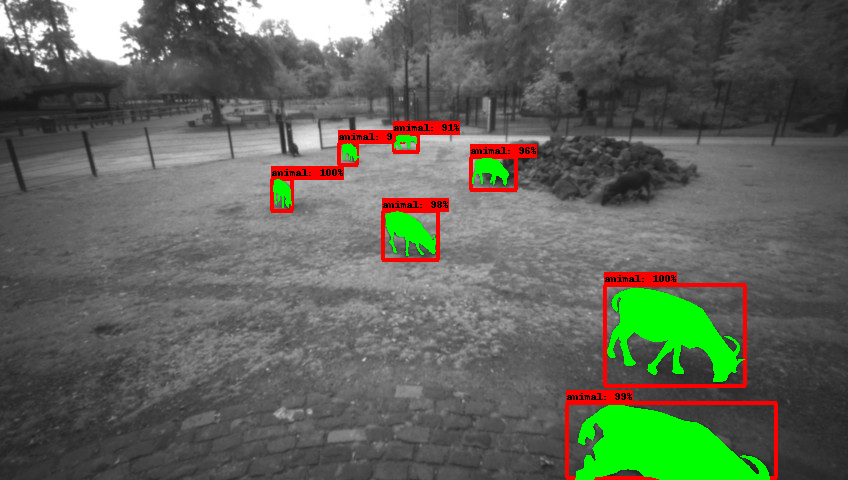}
    \caption{Left: Visual detections of animals depicted solely by bounding boxes. Right: Visual detections of animals depicted by bounding boxes plus segmentation masks.}
    \label{fig:Mask&BB}
\end{figure}
} 
\section{Materials and Methods}

\noindent
\rev{
Our processing pipeline for fully automated distance estimation (\mabbr{}) is based on Deep learning methods. Deep learning methods form a class of machine learning algorithms that have led since 2012 to a breakthrough in computer vision and visual recognition, esp. in the fields of object recognition and detection in images and video clips \citep{AlexNet}.

Since training data is used for the training of machine learning approaches, we first introduce the data material that has been used for training of \mabbr{}. 
Then we explain each module of \mabbr{} and it's functionality.
} 

\subsection{Data Material}\label{sec:data}
\rev{
\noindent
The data material was selected according to the following criteria:
\begin{itemize}
    \item \textbf{Outdoor and wildlife scenarios}. The training data for processing camera-trap imagery should cover image data from as many outdoor and wildlife environments as possible.
    \item \textbf{Absolute depth information}. To train the estimation of metric distances between observed animals and camera-traps, the training data must also show so-called RGB-D imagery. In RGB-D images every image pixel not only shows the color information in terms of it's red, green, and blue color components but also the depth information where the depth value gives the distance between the camera and observed scene part depicted in the pixel.
    \item \textbf{Known field of view}. For the estimation of the real animal-camera distances, \mabbr{} has to create an internal three-dimensional representation of the observed wildlife scenario, a so-called a 3D point-cloud. For this purpose, the opening angle or field of view of the camera-trap must be available in the training data.
\end{itemize}
\noindent
Following these criteria, we settled on the following data material. An overview of their characteristics can be found in Table \ref{datasets_properties}.
} 
\rev{
\begin{itemize}
    \item \textbf{UASOL} \citep{uasol} is a stereo dataset recorded from pedestrian perspectives at the campus of the University of Alicante (Spain). We selected those five scenarios out of 33 available scenarios that contain the most outdoor components and visible vegetation, i.e., the scenarios EPS4, Garden, Nursery, Optics, and Philosophy 1.
    \item \textbf{TartanAir} \citep{tartanair2020iros} is a photo-realistic synthetic dataset captured from persepctives of a flying drone and rendered with Unreal Engine. We decided to use five of the 30 given scenes that were recorded in the outdoor environments which contain the most vegetation: Gascola, Neighborhood, Seasons Forest and Seasons Forest Winter. 
    \item \textbf{DIML} \citep{diml} is a RGB-D-dataset consisting of more than 200 different indoor and outdoor scenes recorded with a Microsoft Kinect V2 and a ZED stereo camera. We decided to use the Scenes Field 1 and Field 2, as these depicted scenes are comparable to camera trap videos. 
    \item \textbf{LVPD} \citep{lvpd} is a forest environment dataset collected in woodland areas in Southampton Common (Hampshire, UK). The camera was mounted 15 cm above the ground on a broom-like contraption to simulate the perspective of a robot ground rover. The images provided by this dataset were most similar to real world camera trap videos of a camera mounted to a tree in a dense forest biosphere. 
    \item \textbf{Lindenthal} is, to our best knowledge, the only outdoor dataset that provides depth as well as tracking information of observed animals \citep{DBLP:journals/corr/abs-2102-05607}. It was recorded by an Intel RealSense D435i stereo camera which was mounted above an animal enclosure at the Lindenthal Zoo (Cologne, Germany). The near infrared camera of the Intel RealSense D435i was used during day- and nighttime to capture gray scale video at 15 frames per second. At nighttime, an infrared lamp was used for active illumination. The animals observed are: geese, goats, donkeys and deer. There is a total amount of 14 scenes, which we enumerated from S00 to S13 (Table \ref{Lindenthal_Overview}). 
\end{itemize}
\noindent
From the more technical viewpoint: The DIML dataset has been used as the validation dataset in training, i.e. for an unbiased tuning of the model hyperparameters. The Lindenthal dataset has been used as test dataset, i.e., for the unbiased final evaluation.
} 
%

\begin{table}[ht]
\resizebox{\textwidth}{!}{%
\begin{tabular}{@{}lllllllll@{}}
\toprule
\textbf{Dataset} & \textbf{Frames} & \textbf{Scenes} & \textbf{Resolution} & \textbf{Acquisition}   & \textbf{Video} & \textbf{Focal Length} & \textbf{HFOV} & \textbf{Max Depth} \\ \midrule
UASOL            & 25.7 K          & 5               & 2208 x 1242         & Stereo ZED             & yes            & 1399.74               & 76.5          & 20                 \\
TartanAir        & 24.5 K          & 5               & 640 x 480           & Rendered                 & yes            & 320                   & 90            & 65                 \\
DIML             & 3 K             & 3               & 1920 x 1080         & Stereo ZED             &                & 1400                  & 69            & 65                 \\
LVPD             & 9.7 K           & 5               & 640 x 480           & Stereo RealSense D435i &                & 462.14                & 74            & 10                 \\
Lindenthal       & 5.8 K           & 14              & 848 x 480           & Stereo RealSense D435  & yes            & 424.74                & 90            & 65                 \\ \bottomrule
\end{tabular}%
}
\caption{Datasets characteristics: Dataset Name, Number of Images, Number of Scenes, Resolution of RGB and Depth-Images, Acquisition of Images, Scenes are Videos, Focal Length [px], Horitzontal Fiel of View [degree], Maximum Depth [m]}
\label{datasets_properties}
\end{table}

\begin{wraptable}{L}{0.5\textwidth}
\centering
\resizebox{0.48\textwidth}{!}{%
\begin{tabular}{@{}llll@{}}
\toprule
\textbf{Scene} & \textbf{Time} & \textbf{Animals} & \textbf{Annotated} \\ \midrule
00             & Day           & Geese, Ducks     & No                 \\
01             & Day           & 9 Goats          & Yes                \\
02             & Day           & 2 Donkeys        & Yes                \\
03             & Day           & 1 Deer           & Yes                \\
04             & Day           & 6 Geese          & Yes                \\
05             & Night         & 2 Deer           & Yes                \\
06             & Evening       & 2 Deer           & Yes                \\
07             & Night         & 1 Deer           & Yes                \\
08             & Day           & None             & No                 \\
09             & Day           & 5 Deer           & Yes                \\
10             & Night         & 2 Deer           & Yes                \\
11             & Night         & 1 Deer           & Yes                \\
12             & Night         & 2 Deer           & Yes                \\
13             & Night         & 1 Deer           & Yes                \\ \bottomrule
\end{tabular}%
}
\caption{Overview over the different scenes in the Lindenthal dataset, enumerated from S00 to S13}
\label{Lindenthal_Overview}
\end{wraptable}

\subsection{\mname{}}
\label{sec:Pipeline}

\begin{figure}
    \centering
    \includegraphics[scale=0.5]{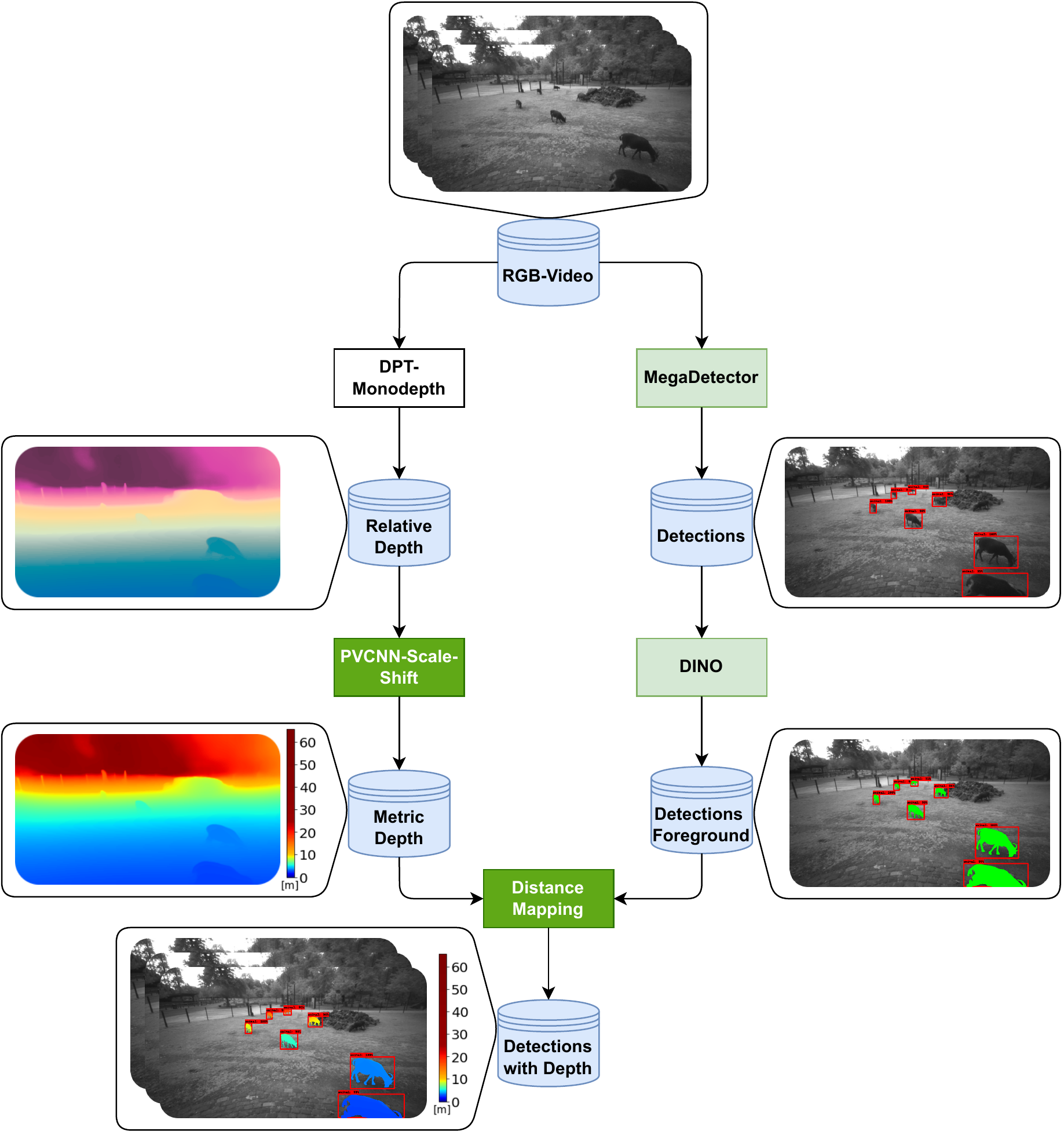}
    \caption{Pipeline Architecture: The input are a RGB video clips, the output consists of the 3D tracks of the observed animals. Newly developed modules are are highlighted in green colour.}
    \label{fig:pipeline_architecture}
\end{figure}

\mabbr{} (Fig. \ref{fig:pipeline_architecture}) takes video clips from conventional camera-traps as input. These video clips can be color video clips taken at daytime or gray-value video clips taken at dawn or nighttime using infrared cameras and infrared illumination. 

\mabbr{} shows two parallel processing branches. The depth estimation branch (left) derives absolute depth information for the complete observed scene, i.e., all observed animals \textit{\textbf{and}} the background (i.e., all visible plants, rocks, trees etc.). The localization branch (right) derives now the missing information, i.e.: where are the visible animals in the observed scene? 

The depth estimation branch (left) first derives the \textbf{\textit{relative depth}} for every video frame using the DPT-Monodepth model (Dense Prediction Transformer, Section \ref{sec:dpt}, \citep{Ranftl2021}). The relative depth information without a metric scale is then aligned by the PVCNN-module (Point-Voxel Convolutional Neural Network, Section \ref{sec:pvcnn}, \citep{liu2019pvcnn}) to to absolute depth information with distance values in meters. 

In the localization branch (right), animals are visually detected in each video frame by using the MegaDetector-framework (Section \ref{sec:megadet}, \citep{beery2019efficient}) that outputs the bounding boxes of every animal detection (red rectangles). For every bounding box, we employ a newly adapted DINO method (self-DIstillation with NO labels, Section \ref{sec:DINO}, \citep{caron2021emerging}) in a multi-instance approach to extract a segmentation mask for every detected animal.

We now have to combine the results of both branches to obtain the desired animal-camera distances. We achieve that by applying the segmentation mask of every animal detection derived in the right branch to the corresponding absolute depth information derived in the left branch and by determining the median of the so selected absolute depth values as the absolute animal-camera distance in meters. 


\subsubsection{Deriving Relative Depth in the Left Branch: DPT}\label{sec:dpt}

For the relative depth estimation, we use the DPT (Dense Prediction Transformer) model developed by \citep{Ranftl2021}. Combined with a large amount of diverse depth datasets, the authors achieve a new state-of-the-art performance during the evaluation on unseen datasets and thus create a robust model for a wide variety of scenes. However, the model only estimates the relative depth and not the absolute metric depth in meters to avoid instability due to the wide range of possible depth scales in the training data.

\textbf{Adapting DPT:} Technically, DPT derives for each input frame a so-called disparity image $\hat{d}$. Such a disparity image encodes the relative depth information by the differences in coordinates of corresponding image points. The values in such a disparity image are inversely proportional to the scene depth at the corresponding pixel location. To obtain in the end depth information, we convert such a disparity image $\hat{d}$ to a first approximation of a depth image $\overline{d}$ (Eq. \ref{eq:approxInv}). We determine the necessary conversion parameters scale $\overline{m} 
$ and shift $\overline{c} 
$ by aligning each DPT disparity output of every image in the training dataset to its disparity Groundtruth via RANSAC (Random Sample Consensus, Section 3.1) and averaging across the resulting scales and shifts.
\begin{equation}\label{eq:approxInv}
   \overline{d} = \frac{1}{\hat{d} \cdot \overline{m} + \overline{c}}
\end{equation}

\subsubsection{Absolute Distance Estimation in the Left Branch: PVCNN} \label{sec:pvcnn}
%
%
Finally, we want to calculate a metric depth estimation for each input image. To this end, we now have to align the approximated depth output of DPT $\overline{d}$ again with a scale $m$ and a shift $c$ to a metric depth image $d_m$, such that
\begin{equation}
    d_m = m \cdot \overline{d} + c
\end{equation}
This time, the scale parameter $m$ determines the visible range of depth values, while the shift parameter $c$ determines the distance of the closest object to the camera and the lowest value of the depth range.

To derive these both parameters, we adapt and modify an approach by \citep{Wei2021CVPR} to recover the 3D shape of an observed scene from just a single image. Thereby, \citep{Wei2021CVPR} estimate a relative depth image, convert it to a point cloud representation (i.e., a set of three-dimensional points where in this case all points origin from the pixels of the approximate depth image), and then utilize a Point-Voxel-CNN (PVCNN) \citep{liu2019pvcnn} to estimate the focal length and the shift needed to create 
three-dimensional reconstruction of an observed scene. 

\textbf{Adapting PVCNN:} Technically, we extend the PVCNN architecture of \citep{liu2019pvcnn} to estimate both scale $m$ and shift $c$, as well as by introducing extensive data augmentation and a novel training regime.

\paragraph{Data Augmentation}\label{sec:data_aug}
Generally, data augmentation techniques are used in machine learning to increase the amount of training data by adding slightly modified copies of already existing data or newly created synthetic data from existing data.
We apply the following augmentation steps to the approximated depth images $\overline{d}$ to improve generalization in training with respect to unseen resolutions, unseen scenes and different focal lengths:
\begin{itemize}
    \item Flipping: Random horizontal flips of given depth images from training data with a probability of 0.5
    \item Cropping: Random crops of given depth images from training data to 16:9 or 4:3 aspect ratio
    \item Scaling: Select random factor $s \in [0.75, 1]$ and multiply centered crop of a given depth image and its depth ground truth in the training data by $s$; then resize this scaled copy to the original resolution and finally multiply the corresponding focal length by $\frac{1}{s}$.
\end{itemize}

\begin{figure}[ht]
    \centering
    \includegraphics[width=\textwidth
    ]{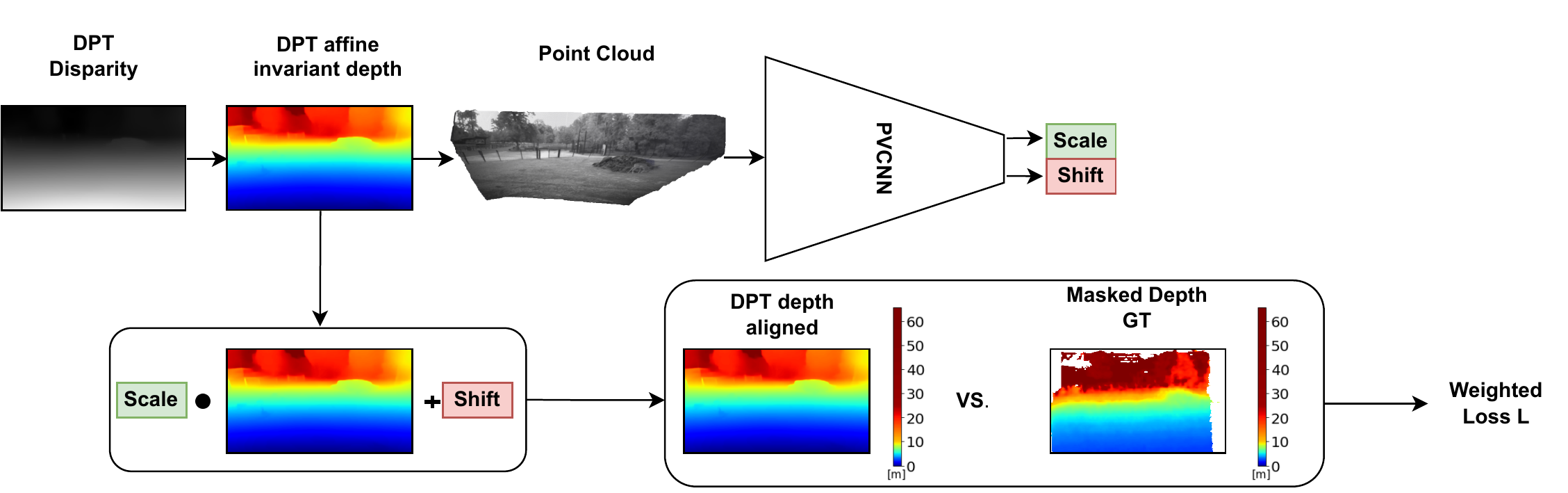}
    \caption{Proposed Training Regime: The DPT disparity image is converted to a point cloud and given to the PVCNN as input. The PVCNN estimates the needed scale and shift, these parameters are applied to the DPT affine invariant depth and then the loss is calculated using the aligned image, ground truth and weighted loss (Eq. \ref{loss}).}
    \label{fig:training_new}
\end{figure}

\paragraph{Training Regime}\label{training_regime}
A new training regime (Fig. \ref{fig:training_new}) shows the following steps: 

First, we unproject the relative depth image back to 3D space similar to \citep{Wei2021CVPR}. In more detail, we assume as camera model a pinhole camera for the point cloud reconstruction and convert 2D image coordinates to 3D by: 
\begin{equation}
\begin{cases}
 &\text{ $x =\frac{u-u_0}{f}\cdot\overline{d}$}\\
 &\text{ $y=\frac{v-v_0}{f}\cdot \overline{d}$} \\
 &\text{ $z = \overline{d}$} \\
\end{cases}
\label{reprojection}
\end{equation}
where $(u_0,v_0)$ is the optical center of the camera, $f$ is the focal length, and $\overline{d}$ is the approximated depth. 

Contrary to \citep{Wei2021CVPR}, we presume that the focal length of the camera is known, since \rev{the focal length of commercial camera traps can usually be found in the specifications provided by the manufacturer or can be easily calculated from the opening angle of the camera (Field Of View, FOV) and the image width in pixels: \begin{equation}
    \text{focal}_\text{pix} = \frac{\text{imgWidth}_\text{pix} \cdot 0.5}{\tan\left( \frac{\text{FOV}_\text{deg} \cdot 0.5 \cdot \pi }{180}\right)}
\end{equation}} 

The point cloud is then given to the PVCNN as input. From there on, the PVCNN estimates the needed scale and shift for the input image. 
In training of a deep learning model, the training loss is a metric used to assess how the model fits the samples of training data that show the correct output (ground truth) for every sample. The training loss is then minimized to improve the model performance.
To calculate this training loss, we align the initial approximated depth input with the scale and shift and apply our loss function (Eq. \ref{loss}) to the output and the ground truth.

Now it is important to note that simply calculating a common pixel-wise loss (difference between derived result and ground truth) would have the disadvantage that training would aim to minimize the loss for all pixels in the same way and would be susceptible to either outliers in the depth estimation of DPT (especially for high distances) or to errors in the ground truth. 

Instead, we propose a weighted loss function that shifts the learning objective to pixels closer to the camera. Let $d_m$ be the aligned metric depth image, $g$ the depth ground truth image, $n_{valid}$ the number of valid pixels in the ground truth, $\exp$ the exponential function, and $\alpha$ the weight factor, then we define the weighted loss $\mathcal{L}_w$ as:
\begin{equation}
    \mathcal{L}_w(d_m,g) = \frac{(d_m-g)^2\cdot \exp(-\alpha \cdot g)}{n_\text{valid}}
\label{loss}
\end{equation}

The factor $\alpha$ controls how much closer pixels with lower depth values influence the overall loss. As a result, we set $\alpha$ to $0.04$ during training, as it achieved the best results on the validation data. 



We train our model for seven epochs on batches of 50 images, 
employ a learning rate of 0.0001 with a decay factor of 0.1 applied every fourth epoch, and we train with a dropout probability of 0.3 for the classifier layer. 

For comparison: Direct training of the scale and shift parameters using the well-established RANSAC method \citep{ransac} yields inferior results during training. Furthermore, we decided against an approach that estimates the two parameters directly from images or image features. As \citep{Wei2021CVPR} observed, ``the domain gap is significantly less of an issue for point clouds than for images'' for this kind of task, which requires an accurate 3D reconstruction of the scene.

\subsubsection{Animal Detection by Bounding Boxes in the Right Branch: MegaDetector}\label{sec:megadet}
The MegaDetector is an animal detection model for camera trap footage proposed by \citep{beery2019efficient} and was trained on several hundred thousand animal detections from camera trap videos recorded in diverse biospheres and of a large variety of animals. We decided for the MegaDetector because of its robustness: It is able to localize animals and species not seen during training and it reliably detects animals in unseen ecosystems and weather conditions as well.

\subsubsection{Animal Detection by Segmentation Masks in the Right Branch: Multi-Instance DINO}\label{sec:DINO}

\begin{figure}[h]
\centering
 \includegraphics[width=\textwidth]{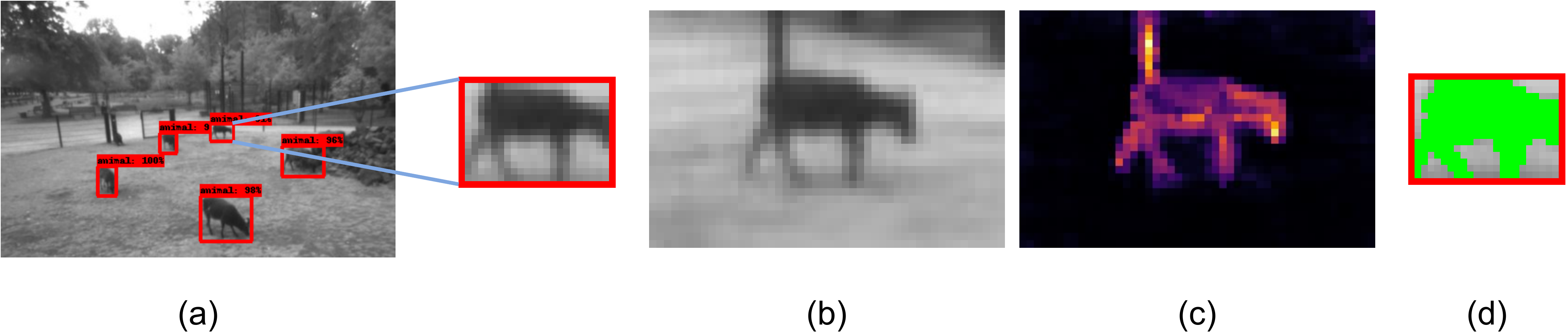}
\caption{Original Detection (a); Wider Region around Detection with more Context (b), DINO Attention Map Output (c), Original Detection with Foreground Segmentation (d) }
\label{DINO_Processing}
\end{figure}

The DINO approach by \citet{caron2021emerging} stands for self-DIstillation with NO labels and describes a method 
to learn class features to classify the detected animals as dear, boar, etc. and so-called attention maps (Fig. \ref{DINO_Processing} (c)). In simple words: an attention map indicates what image locations are important for each animal detection. Thereby, these attention maps can be used to derive the segmentation masks of the detected animals by depicting the pixels belonging to a detected animal inside the bounding boxes (Fig. \ref{DINO_Processing} (d)).
We decided for DINO as the segmentation model because it is an unsupervised machine learning approach, i.e., DINO requires no additional training. Furthermore, DINO correctly handles partial occlusion by vegetation and provides a precise segmentation result. 

The original DINO was trained on the ImageNet dataset \citep{ImageNet}, which mainly contains images showing only one target object (car, truck, cat, etc.) to detect and identify. Therefore, we apply DINO not to the complete camera trap images but instead apply it separately to the bounding box of each animal detected by the MegaDetector.

We call this adapted version \textbf{Multi-Instance DINO} and it shows the following steps (cf. Fig.\ref{DINO_Processing}):
\begin{enumerate}
    \item Input: The bounding box of a detection (a)
    \item Increase the bounding box size by doubling its height and width (b) to mimic the format of the ImageNet dataset and as a consequence create optimal input images for DINO to operate on
    \item Generate the attention map of image crop within this extended bounding box using DINO (c)
    \item Create a segmentation mask by thresholding the attention map at 10\% of the maximum attention value (d)
\end{enumerate}
The segmentation mask is then used to determine the distance of the animal by taking the median of the corresponding depth pixel values in the aligned depth image.
\section{Evaluation and Discussion}

We examine the performance of the distance estimation module by a zero-shot evaluation, i.e., by evaluating the distance estimation module on test samples from a location that were not used during training. For this zero-shot evaluation, we decided for the Lindenthal-dataset, the only outdoor dataset that provides depth as well as tracking information of observed animals \citep{DBLP:journals/corr/abs-2102-05607} (cf. section \ref{sec:data}). The evaluation mirrors the two branches of \mabbr{}, that is, the depth estimation branch and the localization branch.

\subsection{Evaluation of Distance Estimation}\label{ref:MDE_Evaluation}

The proposed distance estimation module consists of two steps: The DPT-based relative distance estimation and the alignment of relative distances to absolute distance estimations via the PVCNN step. Here, it is important to note that the first step of relative distance estimation via DPT is not an original contribution of this proof-of-concept study. A comparative evaluation of DPT is reported by \citet{Ranftl2021}.

Consequently, we do not evaluate the DPT module against other depth estimation approaches. Instead, we juxtapose the following step using the adapted PVCNN module with the Random Sample Consensus (RANSAC) \citep{ransac} alignment method. We evaluate its alignment quality by comparing a transformed DPT-depth image with its corresponding ground-truth image. We consider the complete depth image and the median depth value of the segmentation mask of each detection separately. 

For using RANSAC on every image to align the DPT-based relative disparity $\hat{d}$ (Section 2.2) with the ground truth $g$, we invert $g$ and estimate the unknown scale $m^*$ and unknown shift $c^*$ using RANSAC such that the parameters minimize the absolute disparity error:
\begin{equation}
    (m^*, c^*) \approx \operatorname*{arg\,min}_{m^*, c^*} |m^* \cdot \hat{d} + c^* - \frac{1}{g}|
\end{equation}
Next, we convert our relative disparity $\hat{d}$ to a metric depth image $d$ using $m^*$ and $c^*$:
\begin{equation}
    d = \frac{1}{\hat{d}\cdot m^* + c^*}
\end{equation}

To select one distance value for every detection, the segmentation mask of an animal (Fig. \ref{fig:Mask&BB}) is applied to the corresponding depth image and the median of their values is taken. While RANSAC processes the DPT disparity image and the ground truth depth (Section \ref{sec:data}) for alignment, PVCNN only takes the approximated depth as input.

Afterwards, the resulting aligned depth images of the two approches, i.e., RANSAC-based alignment and PVCNN-based alignment are compared to the ground truth of Lindenthal depth images for distance values smaller than 25 meters, as this is the realistic application range for camera trap videos
\citep{chimpanzee}, \citep{Corlatti}. The animal enclosure observed in the Lindenthal dataset as well as the annotated animals are located in a distance smaller than 20 meters from the camera. 

A training epoch on an Intel Xeon 4215, a Nvidia P5000, and 30 GB of RAM took approximately 2.5 hours, while the loading and augmentation of the ground truth, as well as the precomputed DPT images, were responsible for most of the processing time.\newline

\textbf{Metrics: }Table \ref{metrics} depicts the spatial depth metrics that are commonly applied. $N$ denotes the total number of valid pixels; invalid pixels are masked out during evaluation. $d_i$ and $g_i$ are the estimated and ground truth depths of pixel i, respectively:

\begin{table}[H]
\centering
\resizebox{0.8\textwidth}{!}{%
\begingroup
\renewcommand{\arraystretch}{2}
\begin{tabular}{|ll|ll|}
\hline
RMS: & $\sqrt{\frac{1}{N}\sum_{i=1}^N(d_i-g_i)^2}$ &
MAE: & $\frac{1}{N}\sum^N_{i=1}|d_i-g_i|$ \\
\hline
Rel: & $\frac{1}{N}\sum^N_{i=1}\frac{||d_i - g_i||_1}{g_i}$ &
ME: & $\frac{1}{N}\sum^N_{i=1}(d_i-g_i)$ \\
\hline
\end{tabular}%
\endgroup
}
\caption{Spatial depth metrics: Root mean squared error (RMS), Mean relative error (Rel), Mean Absolute Error (MAE), Mean Error (ME).}
\label{metrics}
\end{table}

\textbf{Results: }Table \ref{lindenthal_depth_metrics} depicts the results of the comparative evaluation of the adapted PVCNN-based alignment method against the RANSAC-based alignment method for animal-camera distances of 25 m maximum. It is important to note that the RANSAC algorithm has the advantage of using the ground truth of the Lindenthal dataset as the alignment goal while the PVCNN-based alignment method has never seen the Lindenthal dataset during training but is only using the relative depth images of DPT. Nevertheless, the PVCNN-based alignment method is not far behind RANSAC in the REL and MAE metrics, that is, only by 19 cm in the MAE and by 0.05 in the REL, while PVCNN outperforms RANSAC with respect to the RMS and ME.

For methods such as CTDS, the accuracy of depth estimation on animal instances is most relevant. Therefore, we additionally evaluate the PVCNN performance on the provided ground truth bounding boxes. For each such bounding box, we apply DINO (Section \ref{sec:DINO}) to separate animal from background pixels and then use the median value of the corresponding depth pixels as an estimation of the animal distance. We compare this value to the ground truth distance of the animal extracted from the depth images with the median of the annotated pixel mask. The metrics display an additional improvement, with an MAE of only 0.99 m, a significantly lower RMS of only 1.68 and a REL of 0.113, suggesting a higher precision of the distance estimation for closer and non-background objects.

\begin{figure}
\centering
\includegraphics[width=0.75\textwidth]{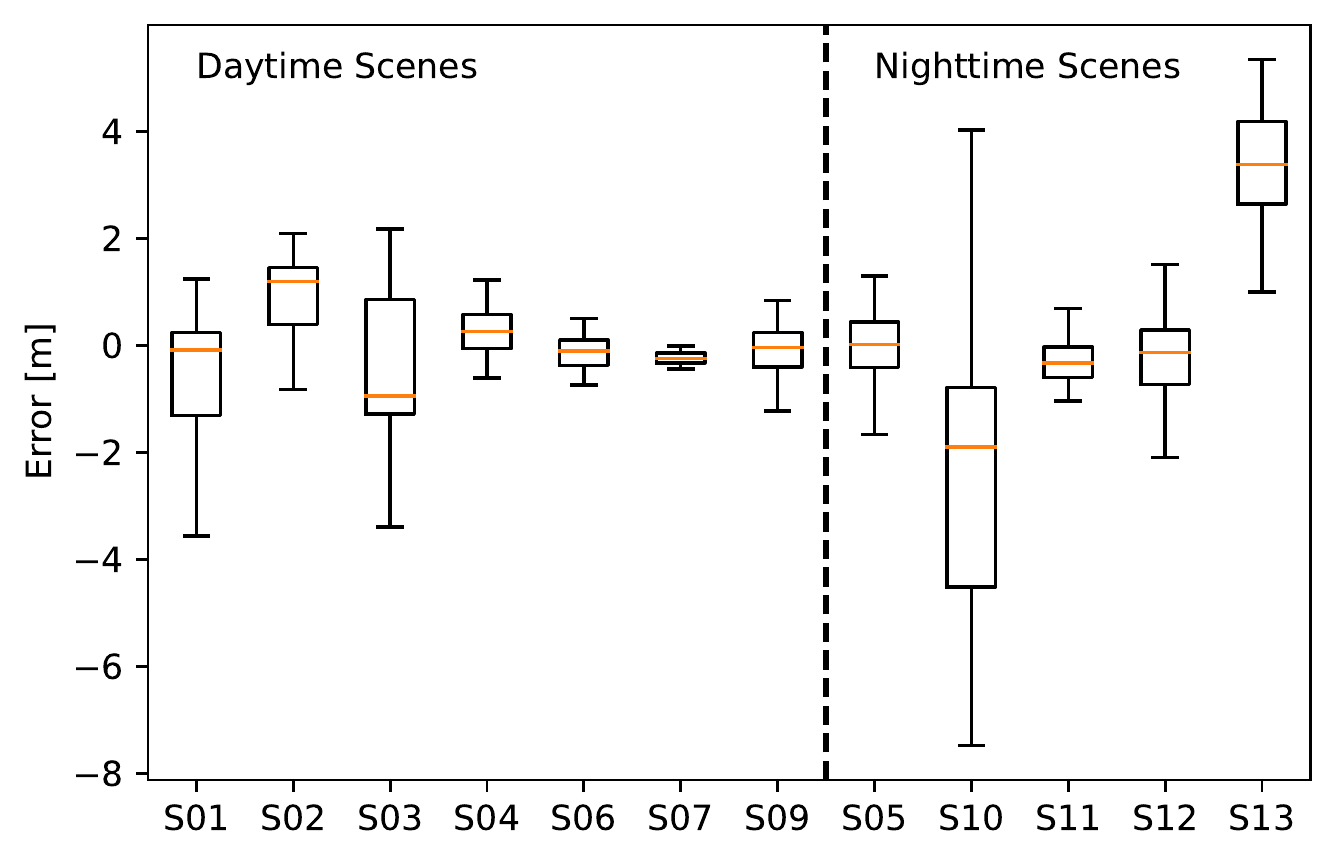}
 \caption{Box plot of the distance estimation error on ground truth annotations per scene up to 25 m. S00 and S08 are not listed here, as no ground truth annotation was available for these scenes.}
\label{fig:Box_Plot}
\end{figure}

\rev{Figure \ref{fig:Box_Plot} further visualizes the distance estimation error averaged over all detected animals for each scene in the Lindenthal dataset (Table \ref{Lindenthal_Overview}). We generally see a low median error (orange line) in most scenes except for S10 and S13. These two scenes show a part of the roof the camera was mounted under at the top of the image. This object introduces a new reference point without any context to the rest of the scene close to the camera, which causes DPT to output highly variant output value ranges. Consequently, this leads to a high spread of estimated alignment parameters and to higher errors (Fig. \ref{fig:Box_Plot}).}

\begin{table}[H]
\resizebox{\textwidth}{!}{%
\begin{tabular}{llllll}
\hline

                                                                                   &                                            & \textbf{RMS$\downarrow$} & \textbf{REL$\downarrow$} & \textbf{MAE$\downarrow$} & \textbf{ME$\downarrow$} \\ \hline
\multicolumn{1}{l|}{Complete GT image}                                             & \multicolumn{1}{l|}{PVCNN}             & 2.5695                   & 0.1428                   & 1.1978                  & -0.2322                 \\
\multicolumn{1}{l|}{}                                                              & \multicolumn{1}{l|}{RANSAC}                & 2.8772                  & 0.0985                   & 1.0075                   & 0.4411                 \\ \hline
\multicolumn{1}{l|}{Instance Depth on GT BBs}                                      & \multicolumn{1}{l|}{PVCNN + DINO + Median} & 1.6821                   & 0.1130                   & 0.9864                   & 0.1754                  \\
\end{tabular}%
}
\caption{Comparative evaluation results using RANSAC and PVCNN for distance estimation using the depth metrics of table \ref{metrics}.}
\label{lindenthal_depth_metrics}
\end{table}

\rev{%

\textbf{Comparison with other automatic methods:} In the \textit{DeepChimpact} competition, organized by DrivenData \citep{drivendata}, the goal was to estimate animal distance from camera trap images. Training and testing was performed on mutually exclusive subsets of a single dataset. In other words, no zero-shot evaluation was used, and the models might first need to be re-trained if applied to new datasets. The winning entry achieved a mean absolute error of 1.6203 m \citep{deep_chimpact_results}. Another semi-automated approach achieves a mean absolute error of 1.8527 m \citep{haucke2022overcoming}.

\textbf{Comparison with manual distance estimations:} Traditionally, distance estimations in ecology have been carried out by humans. Here, distances are not estimated in a continuous fashion, but instead assigned to intervals of at least 1 m. For example, \citet{ctds} assign animals to 1 m intervals out to 8 m, and then increase the interval size for larger distances. This illustrates that the resulting manual distance estimations are inherently coarse. They are also not objective, as shown in the user study of \citet{haucke2022overcoming}. This study resulted in a mean standard deviation between five participants of 0.62 m, a pairwise MAE of 0.7796 m, and a mean relative error of 0.2189 m. In comparison, our method achieves a mean absolute error of $\text{MAE}_\text{instance}=\SI{0.9864}{\meter}$, a mean error (bias) of $\text{ME}_\text{instance}=\SI{0.1754}{\meter}$ and a mean relative error of $\text{REL}_\text{instance}=\SI{0.1130}{\meter}$ (Table \ref{lindenthal_depth_metrics}). Errors may be influenced by factors such as the distribution of distances present in the image (errors tend to get larger with growing distance) and animal visibility (the distance of poorly visible animals is harder to estimate). As we evaluate our method on the novel Lindenthal dataset, some of these factors might influence the above comparison. However, the lower mean relative error suggests that our method is overall more accurate at larger distances than the participants in the user study conducted by \citet{haucke2022overcoming}.}%

\subsection{Degree of Automation}
\begin{figure}
    \centering
    \includegraphics[width=\textwidth]{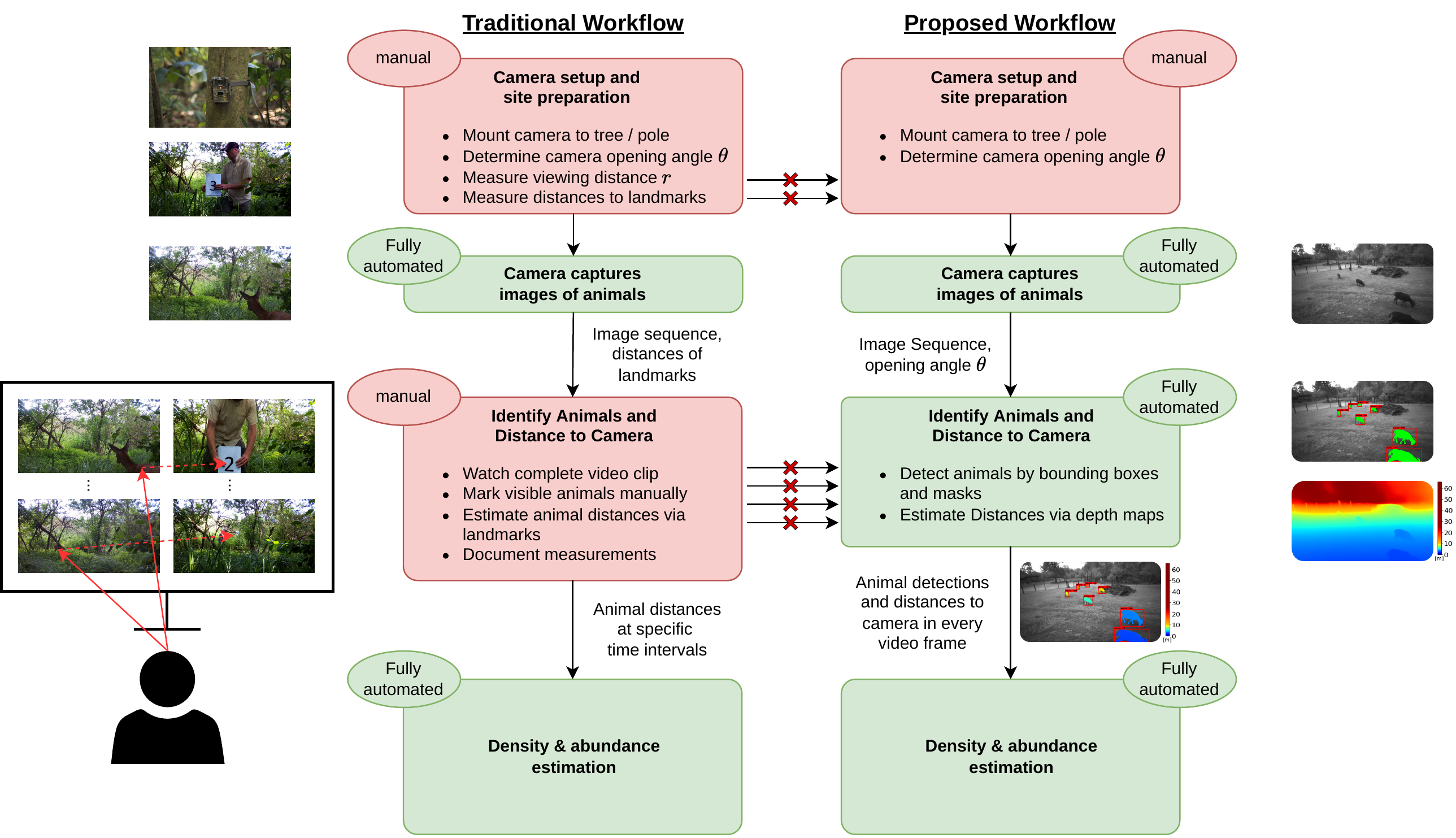}
    \caption{\rev{Comparison traditional workflow (left) vs. improved workflow with \mabbr{} (right). Crossed out arrows designate steps which are no longer needed in our approach. The traditional workflow needs reference footage, e.g. in the form of paper sheets designating the respective distance, as shown here. To estimate distances, animals must be localized (solid red arrows) and their position manually matched with the reference material (dashed red arrows). In contrast, our \mabbr{} fully automates the animal localization and distance estimation.}}
    \label{fig:workflow}
\end{figure}

\rev{We compare the traditional workflow and our novel \mabbr{} in figure \ref{fig:workflow}. The traditional workflow requires capturing reference footage, e.g. by placing a measuring tape in the scene and then holding up a paper sign with the respective distance in 1 m intervals. In contrast, our method does not need any reference footage, significantly reducing the effort required during the camera setup. In the next step of the traditional workflow, researchers will need to (1) watch the observation videos, (2) localize animals appearing in the video, (3) compare the animal locations with reference material to obtain a distance estimation, and (4) document the measurement. This process takes an experienced individual roughly 10 minutes per 1 minute of video \citep{kuehl22PersCom}. In contrast, our method is fully automatic and only requires images or video depicting animals together with the focal length specification of the corresponding camera trap. On a computer with an Intel Xeon 4215 CPU, 30 GB of RAM and an Nvidia P5000 GPU, our method takes about 0.5 seconds per image / video frame to estimate animal distances. This process can be left unattended. By saving this manual effort, the complete automation of the process enables the possibility of large-scale animal abundance studies and could hence accelerate biodiversity research.}

\subsection{Applicability and Application Potentials}\label{sec:SORT}

\rev{Distance sampling relies much more on low bias than the magnitude of random errors \citep{buckland2004advanced}. As our bias is relatively low ($\text{ME}_\text{instance}=\SI{0.1754}{\meter}$, we argue that our distance estimations are well-suited for CTDS in real-world scenarios. While our proof-of-concept study focuses on the application to CTDS, we conjecture that our methodology transcends to other abundance estimation methods such as the random encounter model \citep{rowcliffe2008estimating}, the random encounter and staying time model \citep{nakashima2018estimating}, the time-to-event model, space-to-event model, and instantaneous estimator \citep{moeller2018three}. This is because the estimation of detection probability is required for all mentioned methods.}

As a challenging example, we demonstrate the application of \mabbr{} for visual tracking of animals in video clips captured by camera traps, which is important for behavioral studies of individual animals and animal herds based on their movements and actions \citep{SCHINDLER2021101215}. Additionally, in the context of this study, reliable tracking is important for one other approach to abundance estimation, namely the Random Encounter Model \citep{rowcliffe2008estimating}, which requires velocity estimations of the observed animals.

For this demonstration, we decided for the SORT (for Simple Online and Realtime Tracking) approach proposed by \citet{Bewley2016_sort}. SORT takes bounding box of animal detections (as derived in the localization branch of \mabbr{}) as input. SORT connects these animal detections over all frames to cohesive tracks based on a Kalman-Filter framework \citep{kalman} and the association metric of two visual detections by Intersection over Union (IoU) \citep{iou}. This means, IoU measures how appropriate the bounding box of an animal detection in a frame fits to a bounding box of an animal detection in the previous frame for continuing the tracking of this detected animal.  

We adapted SORT to include the depth information in the Kalman-Filter predictions and replace $IoU$ with a new customized association metric $SimScore$ (\ref{SimScore}) that combines the traditional $IoU$ with a distance similarity metric $DIST_Z$. $\alpha$ controls the weight of each metric. $DIST_Z$ depends on the hyperparameter $DIST_{max}$. If the depth distance between the tracker prediction $z_T$ and the detection $z_{DET}$ is larger than $DIST_{max}$, $DIST_Z$ is clipped to zero, otherwise the difference is subtracted from $DIST_{max}$ and then normalized. Going forward, we will refer to the adapted SORT version as SORT 2.5D (due to adding and processing depth information).

\begin{equation}
    SimScore = \alpha \cdot IoU + (1-\alpha) \cdot DIST_Z,\; \; \alpha \in [0,1] \newline
\label{SimScore}
\end{equation}
\begin{equation}
    DIST_Z = (\frac{DIST_{max} - |z_{T}-z_{DET}|}{DIST_{max}}).clip(0,1)
\end{equation}

The evaluation employs two established multi object tracking metrics that were developed for the KITTI dataset benchmark: the CLEAR MOT metrics \citep{bernardin2008evaluating}:

Multi Object Tracking Accuracy (MOTA) and Multi Object Tracking Precision are original MOT metrics defined by
\[\text{MOTA} = 1 - \dfrac{\text{FN}+\text{FP}+\text{IDS}}{num_{gt}} \text{ and MOTP} = \frac{\sum_i \text{IoU}_{3D}(D_i^t,B_{\text{pred}}^t)}{TP},\] 
where FN, FP, TP are the false negatives, false positives and true positives, with IDS being the number of identity switches of predicted tracks. A detection is considered to be a true positive if the distance in 3D space to the corresponding ground truth track is smaller than 2.2 meters and to be a false positive if it is higher than 2.2 meters. We decided to use this approximated threshold, as in traditional CTDS via reference images the distance measurements are assigned to intervals of, for example, 1 meter for 0-8 meters, to intervals of 2 meters for 8-12 meters and to an interval of 3 meters for 12-15 meters \citep{chimpanzee}.
While MOTA relies entirely on the fraction of correctly identified individuals, MOTP quantifies the precision of ground truth bounding boxes against predicted bounding boxes. 

\begin{figure}[ht]
\centering
   \begin{minipage}[b]{0.75\textwidth}
   \vspace{1cm}
    \includemedia[
    width=0.9\textwidth,
    height=0.81026393\textwidth,
    activate=pageopen,
    passcontext,
    addresource=Graphics/3d_track_vis/tracked_live_cropped.mp4,
    flashvars={source=Graphics/3d_track_vis/tracked_live_cropped.mp4&loop=true&autoPlay=true}
    ]{\includegraphics[width=0.9\textwidth]{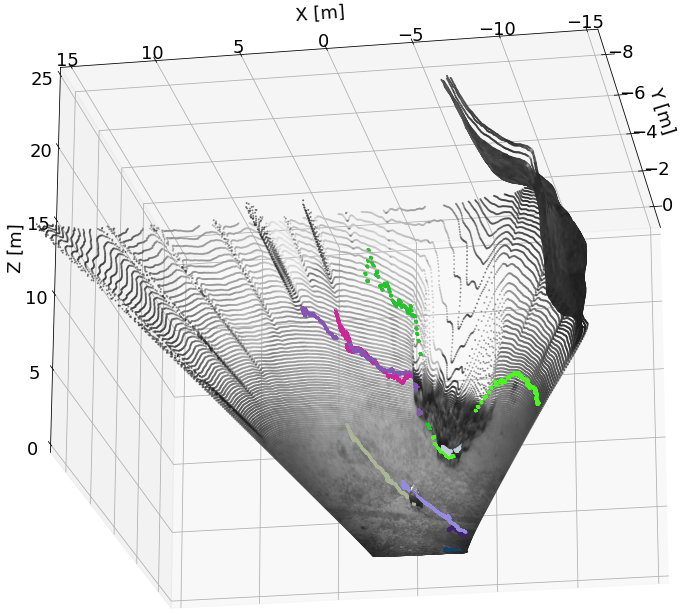}}{VPlayer.swf}
  \end{minipage}
\caption{Qualitative tracking result depicted by 3D point cloud unprojected from depth images of one Lindenthal video clip. In the digital version, compatible PDF readers will allow video playback.}
\label{S01_3D_tracks}
\end{figure}

The SORT 2.5D version achieves a MOTA score of 56.3\%, an average localization precision for correct detections of only 0.648 meters (MOTP) and a high precision of 90.3\%. Figure \ref{S01_3D_tracks} demonstrates a qualitative tracking result.

\section{Conclusion}
\rev{%
We propose \mname{}, a fully automated processing pipeline for estimating animal distances in video and still images of camera traps.
We derive absolute distances in metric values based on monocular relative depth estimation by exploiting a novel 3D point cloud-based alignment model that is trained on a diverse collection of outdoor datasets and thus entirely eliminates the need for reference images. We detect and localize animals using our multi-instance DINO method. We evaluate the optimized approach in a zero-shot evaluation on the Lindenthal zoo scenario dataset, which was not seen during training. On the Lindenthal dataset, we achieve a mean absolute error over all animal instances of only 0.9864 meters and a mean relative error of 0.113. In contrast, the previous automated approaches have much higher mean absolute errors of 1.8527 m \citep{haucke2022overcoming} and 1.6203 m \citep{deep_chimpact_results}. Manual estimations in the user study by \citet{haucke2022overcoming} have a higher mean relative error of 0.2189.
By comparing \mabbr{} with the traditional workflow (Fig. \ref{fig:workflow}), we show that we relieve ecologists of a significant workload, by requiring neither the time-cosuming comparison of observation and reference material nor the capture of any reference material in the first place.
Although we focused on the application to CTDS, we conjecture that our methodology transcends to other abundance estimation methods such as the random encounter model \citep{rowcliffe2008estimating}, the random encounter and staying time model \citep{nakashima2018estimating}, the time-to-event model, space-to-event model, and instantaneous estimator \citep{moeller2018three}.}
\clearpage
\noindent 
\section*{\centering \begin{normalsize}Acknowledgement\end{normalsize}}
This work is partially funded by the German Federal Ministry of Education and Research (Bundesministerium
für Bildung und Forschung (BMBF), Bonn, Gemany (AMMOD - Automated Multisensor Stations for Monitoring of BioDiversity: FKZ 
01LC1903B). This funding is gratefully acknowledged.
\rev{We thank Hjalmar S. Kühl (Max Planck Institute for Evolutionary Anthropology in Leipzig, Germany), Maik Henrich (University of Freiburg \& Bavarian Forest National Park, Germany) and Emeline Auda (Wildlife Conservation Society Cambodia) for fruitful discussions on camera trapping and abundance estimation.

We thank Frank Schindler and Erika Haucke for proofreading the manuscript.}

We thank Thomas Ensch, Michael Gehlen and the entire team of the Lindenthaler Tierpark for their cooperation by hosting the experimental camera trap hardware on-site. We thank Alejandro Berni Garcia for his help with the construction of the wooden camera trap casing.



\clearpage

%
%





\clearpage

\bibliographystyle{apalike} 
\bibliography{references}

\end{document}